\PassOptionsToPackage{table,dvipsnames,svgnames}{xcolor}
\documentclass{article}
\usepackage{PRIMEarxiv}

\usepackage[utf8]{inputenc}
\usepackage[T1]{fontenc}
\usepackage{hyperref}
\usepackage{url}
\usepackage{natbib}
\usepackage{booktabs}
\usepackage{amsfonts}
\usepackage{amssymb}
\usepackage{amsmath}
\usepackage{nicefrac}
\usepackage{microtype}
\usepackage{graphicx}
\usepackage{multirow}
\usepackage{array}
\usepackage{xcolor}
\usepackage{listings}
\usepackage{pifont}
\usepackage{enumitem}
\usepackage{caption}
\usepackage{subcaption}
\usepackage{tabularx}
\usepackage{colortbl}
\usepackage{algorithm}
\usepackage{algpseudocode}
\usepackage{booktabs}
\usepackage{graphicx}
\usepackage{amsmath}
\usepackage{xcolor}

\usepackage{tikz}
\usetikzlibrary{arrows.meta,positioning,fit,backgrounds,calc,shapes.geometric}
\usepackage{helvet} 

\usepackage{tcolorbox}
\tcbuselibrary{skins}
\newtcolorbox{takeaway}[1][Takeaway]{%
  colback=black!4!white,
  colframe=black!60!white,
  fonttitle=\bfseries\small,
  colbacktitle=black!80!white,
  coltitle=white,
  enhanced,
  attach boxed title to top left={yshift=-2.2mm, xshift=4mm},
  boxrule=0.6pt,
  arc=1.5pt,
  left=6pt, right=6pt, top=6pt, bottom=4pt,
  title={#1},
}

\definecolor{darkgreen}{rgb}{0.0,0.42,0.0}
\definecolor{codegreen}{rgb}{0,0.6,0}
\definecolor{codegray}{rgb}{0.5,0.5,0.5}
\definecolor{codepurple}{rgb}{0.58,0,0.82}
\definecolor{backcolour}{rgb}{0.97,0.97,0.97}
\definecolor{stRecover}{RGB}{31,119,180}
\definecolor{stHarden}{RGB}{214,39,40}
\definecolor{stUnlearn}{RGB}{44,160,44}
\definecolor{stSteer}{RGB}{148,103,189}
\definecolor{stBox}{RGB}{238,242,248}
\definecolor{stEval}{RGB}{255,243,224}
\definecolor{stTodo}{RGB}{255,140,0}
\definecolor{taskbg}{RGB}{232, 230, 245}
\definecolor{taskbr}{RGB}{152, 138, 195}
\definecolor{instbg}{RGB}{239, 238, 234}
\definecolor{instbr}{RGB}{176, 176, 168}
\definecolor{ftbg}{RGB}{221, 235, 247}
\definecolor{ftbr}{RGB}{136, 169, 216}
\definecolor{compbg}{RGB}{246, 233, 233}
\definecolor{compbr}{RGB}{183, 112, 111}
\definecolor{hardenC}{RGB}{54, 84, 217}
\definecolor{recoverC}{RGB}{72, 147, 129}
\definecolor{unlearnC}{RGB}{208, 134, 44}
\definecolor{steerC}{RGB}{115, 65, 214}
\definecolor{interpC}{RGB}{102, 113, 132}
\definecolor{evalC}{RGB}{67, 121, 70}
\definecolor{hardenP}{RGB}{234, 240, 255}
\definecolor{recoverP}{RGB}{230, 243, 238}
\definecolor{unlearnP}{RGB}{253, 240, 225}
\definecolor{steerP}{RGB}{242, 233, 255}
\definecolor{grayP}{RGB}{240, 242, 245}

\definecolor{restoredC}{RGB}{240, 255, 240}
\definecolor{restoredBr}{RGB}{76, 175, 80}

\newcommand{\safetune}{\emph{SafeTune}}
\newcommand{\cmark}{\textcolor{darkgreen}{\ding{51}}}
\newcommand{\xmark}{\textcolor{red}{\ding{55}}}
\newcommand{\rr}{\ensuremath{\overline{\mathrm{RR}}}}
\newcommand{\tsafe}{\ensuremath{\tau_{\mathrm{safe}}}}
\newcommand{\RECOVER}{\textsc{Recover}}
\newcommand{\HARDEN}{\textsc{Harden}}
\newcommand{\UNLEARN}{\textsc{Unlearn}}
\newcommand{\STEER}{\textsc{Steer}}

\definecolor{lstbg}{HTML}{F8FAFC}
\definecolor{lstkeyword}{HTML}{1E40AF}
\definecolor{lststring}{HTML}{0D9488}
\definecolor{lstcomment}{HTML}{64748B}
\definecolor{lstfunc}{HTML}{9333EA}
\definecolor{lstrule}{HTML}{CBD5E1}
\definecolor{termTeal}{HTML}{0D9488}
\definecolor{termBlue}{HTML}{1E40AF}
\definecolor{termViolet}{HTML}{9333EA}
\definecolor{termAmber}{HTML}{B45309}
\definecolor{lsttitlebg}{HTML}{1E293B}
\lstdefinelanguage{yaml}{
  morekeywords={true,false,null,y,n},
  sensitive=false,
  morecomment=[l]{\#},
  morestring=[b]",
  morestring=[b]',
  alsoletter={-},
  moredelim=[l][\color{black}\ttfamily]{-},
}
\title{\includegraphics[width=0.81\linewidth]{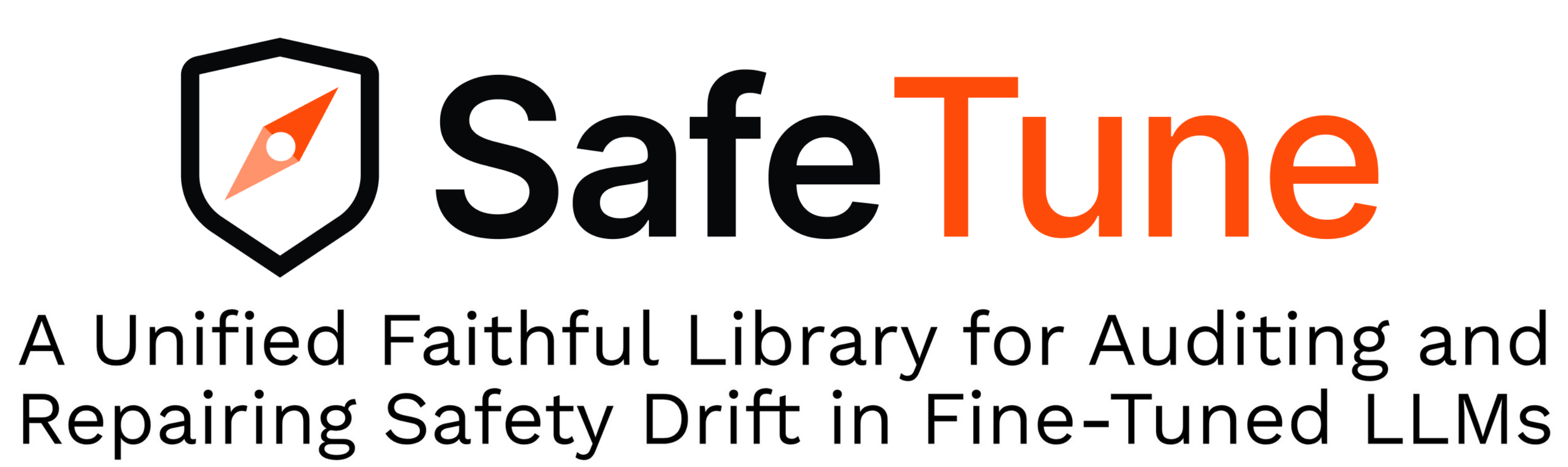}}

\author{
  Pratinav Seth\textsuperscript{*}, Saisab Sadhu\textsuperscript{*}, Anshul Kaushal\textsuperscript{*\dag},\\ Vinay Kumar Sankarapu \\
  \affiliation{Lexsi Labs} \\
  \texttt{pratinav.seth@lexsi.ai}
}

\setabstract{
Methods for addressing safety drift in fine-tuned Large Language Models (LLMs) are scattered across incompatible implementations, lifecycle stages, and evaluation protocols, making them difficult to adopt and compare. We introduce \safetune{}, a source-available library that unifies four intervention paradigms: post-hoc weight recovery, safety-constrained fine-tuning, gradient-based unlearning, and inference-time steering, alongside shared interpretability, evaluation, and deployment utilities. \safetune{} provides a consistent configuration-driven workflow while preserving the distinct inputs and intervention points each paradigm requires. Its modular registry supports new methods, benchmarks, judges, models, and fine-tuning domains without redesigning the surrounding pipeline. We demonstrate \safetune{} through controlled comparisons and finance and medical deployment case studies, showing how it characterizes safety drift, evaluates feasible interventions on common refusal-behavior and capability evaluations, and supports calibrated or layered mitigation. The library and documentation are released at: \url{https://github.com/Lexsi-Labs/SafeTune}.
}

\setkeywords{AI Safety, Large Language Models, Fine-Tuning, Safety Alignment, Source-Available Library, Reproducibility}

\runningtitle{SafeTune: A Unified Faithful Library for Auditing and Repairing Safety Drift in Fine-Tuned LLMs}

\begin{document}
\maketitle


{\renewcommand{\thefootnote}{\fnsymbol{footnote}}\footnotetext[1]{\vspace{0.5em} Equal contribution}\footnotetext[2]{Work done while at Lexsi Labs.}}


\section{Introduction}
\label{sec:intro}

Deploying a large language model typically involves multiple stages
of post-training. The first, safety-aware tuning such as Reinforcement
Learning from Human Feedback (RLHF) or Direct Preference Optimization
(DPO) on human-preference data, establishes refusal behavior
associated with low-dimensional activation and parameter
structures~\citep{zhang2026understanding,arditi2024refusal}. The second,
task-specific fine-tuning on benign downstream data such as medical,
legal, or coding corpora, steers the model toward the new domain and can weaken this refusal
behavior even when the fine-tuning data contains no harmful
examples~\citep{qi2024finetuning,yang2024shadow,zhan2024removing,khan2026safetydrift}.
The resulting checkpoint complies with requests its base model
refused. Because the drift lives in the deployed checkpoint rather
than the released base, the developer who performed the fine-tuning is typically best positioned to measure or repair it.

A repair literature has grown to meet this risk. It spans four
paradigms: post-hoc weight patching, safety-constrained retraining,
gradient-based unlearning, and inference-time steering, together with
more than thirty methods. It is, however, \emph{fragmented
at the level of evaluation}: a typical method is tested on its own
drifted checkpoint, with its own repair implementation, scored by its
own judge, and compared against its own baselines. Existing
tools do not close this gap. Evaluation harnesses such as
HarmBench~\citep{mazeika2024harmbench} and
WildGuard~\citep{han2024wildguard} lack drift-induction and repair
implementations, while comparisons such as
GuardSpace~\citep{zhang2025guardspace}, CWAC~\citep{peng2026cwac}, and
Antibody~\citep{nguyen2026antibody} each leave a variable, paradigm,
checkpoint, or model family uncontrolled (Section~\ref{sec:related}).
A practitioner holding a freshly drifted model and asking ``what
should I run?'' must currently interpret results obtained under substantially different checkpoints, implementations, and evaluation protocols.

\emph{The missing piece is a substrate, not a method.}
Continued development of repair methods must be complemented by a
common substrate on which existing methods can be implemented and
compared under identical conditions. Such a substrate
must do three things that no current tool does jointly.
It must provide reproducible drift-induction recipes that yield
characterized checkpoints of known severity, while accepting
externally produced checkpoints. It must bring all four intervention paradigms into a shared workflow
while preserving their distinct inputs and lifecycle stages.
Implementations should also document their correspondence with the
originating descriptions and any known deviations.

We present \safetune{}, a source-available library that standardizes
the safety-repair pipeline around four components, all reached through
a single configuration-driven Application Programming Interface (API)
(Figure~\ref{fig:arch}, Section~\ref{sec:arch}). A \emph{drift-induction} stage, a fixed and
logged fine-tuning recipe, produces characterized checkpoints with
reproducible severity. An \emph{intervention registry} organizes \RECOVER{} (patching),
\HARDEN{} (constrained retraining), \UNLEARN{} (forgetting), and
\STEER{} (inference-time editing) within a shared construct, execute,
and evaluate pattern. An \emph{evaluation} harness integrates an
extensible suite of safety benchmarks (seven by default) with matched
capability anchors, alongside a read-only instrumentation layer
(\emph{Interpret} and \emph{Evaluate}).


The drift recipe and the intervention guide are reproducibility artifacts
rather than API entry points: the installed API is agnostic to how a
supplied checkpoint came to be drifted. \safetune{} serves practitioners evaluating fine-tuned models before
deployment and researchers implementing or comparing safety
interventions. By varying models, domains, and interventions while
retaining a shared evaluation protocol, the library supports
comparisons that would otherwise require separate pipelines. We
demonstrate this functionality through controlled comparisons and
finance and medical case studies.

\paragraph{Contributions.}
\begin{enumerate}[leftmargin=1.4em,itemsep=2pt,topsep=2pt]
  \item \emph{Unified safety-intervention library.} We introduce
  \safetune{}, a source-available library spanning post-hoc recovery,
  safety-constrained fine-tuning, unlearning, inference-time steering,
  interpretability, evaluation, and deployment guardrails
  (Sections~\ref{sec:design} to~\ref{sec:api}).
  \item \emph{Common workflow across intervention stages.} \safetune{}
  provides a consistent construct, execute, and evaluate pattern while
  preserving the different inputs and outputs required by
  interventions applied during training, after fine-tuning, or at
  inference. Implementations are reviewed against their originating
  descriptions, with corrections and known deviations recorded in the
  repository documentation.
  \item \emph{Extensible evaluation and deployment substrate.} New
  methods, benchmarks, judges, model families, and domains are added
  through modular registries, with shared configuration, evaluation,
  logging, and serving support. We release
  the library, documentation, examples, and configurations, with
  checkpoints and reference models released progressively as
  supporting artifacts.\footnote{Model artifacts will be released soon at \url{https://huggingface.co/collections/Lexsi/safetune-artifacts} .}
\end{enumerate}

\section{Background and Related Work}
\label{sec:related}

\subsection{Benign fine-tuning erodes alignment}
\cite{qi2024finetuning} first showed that fine-tuning a frontier model on a standard instruction corpus, with no harmful examples, measurably raises harmful compliance. The effect replicates across open-weight model families and Low-Rank Adaptation (LoRA) settings~\citep{yang2024shadow,zhan2024removing,lermen2023loraundo,hsiung2025guardrails,kalajdzievski2024forgetting}.

Mechanistically, safety concentrates in a low-rank weight subspace\footnote{$\theta \in \mathbb{R}^d$ for the model parameters, where $d$ is the total parameter count. We use this notation consistently in later sections.} ($\theta$) that broad task gradients overwrite \citep{zhang2026understanding,wei2024brittleness}, with a corresponding refusal direction in the residual stream whose degradation tracks the safety loss \citep{arditi2024refusal}. This low-rank account is contested. \cite{ponkshe2025subspaces} find that safety and capability updates are not linearly separable, so subspace-restricted repair rests on a premise \safetune{} lets a practitioner test rather than assume. \cite{khan2026safetydrift} extend this picture to deployment, auditing publicly released checkpoints and finding benchmark-level heterogeneity \emph{within} the same model family. This heterogeneity shows that a single safety benchmark is an unreliable summary of true safety~\citep{schwinn2026coinflip}, motivating the multi-benchmark protocol that \safetune{} adopts.

\subsection{The four intervention paradigms}
We group the repair literature~\citep{huang2024hftsurvey} into four paradigms:

\emph{(I) Post-hoc weight patching} (\RECOVER{}) applies closed-form arithmetic to a pre-computed safety task vector $\tsafe = \theta_{\mathrm{aligned}} - \theta_{\mathrm{base}}$, where $\theta_{\mathrm{aligned}}$ corresponds to a model aligned through RLHF \citep{ilharco2023task,yang2024shadow,bhardwaj2024resta,zhou2025lssf,djuhera2026safemerge,lu2025safedelta,kasliwal2026ctheta,hazra2024safetyarith,hammoud2024mergesafety,wu2024wheat,zhang2026oneshot}.

\emph{(II) Safety-constrained retraining} (\HARDEN{}) adds an alignment-preservation term to the fine-tuning loss, projecting task gradients off the safety subspace or anchoring to an aligned reference \citep{yi2025safegrad,huang2024vaccine,huang2024lisasafety,huang2025booster,liu2026lookahead,tamirisa2024tar,yuan2024derta,yang2025asft,peng2024safetybasin,chen2025vulnaware,hu2025bayesian}.

\emph{(III) Gradient-based unlearning} (\UNLEARN{}) applies a forgetting objective directly against harmful content; Negative Preference Optimization (NPO) is the canonical instance \citep{zhang2024npo}.

\emph{(IV) Inference-time steering} (\STEER{}) leaves the weights unchanged and edits the residual stream at generation time \citep{arditi2024refusal,panickssery2024caa,turner2023actadd,xu2024safedecoding,wollschlager2025refusalgeometry,zhao2025harmfulness,shen2024jailbreakantidote,obrien2024saerefusal,chawla2026alignbeam}, together with a linear-probe safety-routing baseline, a standard technique with no single originating paper.

\begin{table}[!ht]
\centering
\caption{How to use each paradigm, and what distinguishes the methods
inside it. The calling pattern is uniform; the contract is not, and
the \emph{You supply} column is usually what rules a paradigm out. Calls
are on the paradigm's own namespace. Methods named are representative
positions on each axis, not the full catalogue
(Table~\ref{tab:catalogue}).}
\label{tab:contracts}
\footnotesize
\begin{tabularx}{\textwidth}{@{}l
  >{\hsize=1.42\hsize\linewidth=\hsize\raggedright\arraybackslash}X
  >{\hsize=0.74\hsize\linewidth=\hsize\raggedright\arraybackslash}X
  >{\hsize=0.84\hsize\linewidth=\hsize\raggedright\arraybackslash}X@{}}
\toprule
\textbf{Paradigm} & \textbf{Axis of variation} & \textbf{You supply} & \textbf{Call $\rightarrow$ result} \\
\midrule
\textcolor{stRecover}{\RECOVER} &
\emph{How much of the model the edit touches.} Whole-model
(Task Arithmetic) $\rightarrow$ subspace (LoX) $\rightarrow$ layer
(SafeMerge) $\rightarrow$ neuron (PKE) $\rightarrow$ circuit
(C-$\Delta\Theta$). The more targeted the edit, the more it depends on
locating safety correctly. &
drifted checkpoint $+$ aligned/base references &
\texttt{.apply()}: no optimizer $\rightarrow$ patched checkpoint in
seconds \\
\addlinespace
\textcolor{stHarden}{\HARDEN} &
\emph{How the safety signal enters the fine-tune.} Gradient surgery
(SafeGrad), weight-space regularization (AsFT), representation
perturbation (Vaccine), data shaping (LISA), tamper-resistance (TAR). &
base model $+$ \emph{your} fine-tuning corpus &
\texttt{.train(ds)}: replaces your SFT trainer $\rightarrow$
defended checkpoint, hours of training \\
\addlinespace
\textcolor{stUnlearn}{\UNLEARN} &
\emph{What the forgetting objective optimizes against.} Negated LM
loss (GradientAscent), negative preference (NPO), activation redirect
(RMU), $f$-divergence (FLAT). &
finished model $+$ forget and retain sets &
\texttt{.unlearn(...)}: trains $\rightarrow$ unlearned checkpoint \\
\addlinespace
\textcolor{stSteer}{\STEER} &
\emph{Where the edit lands.} Residual stream (CAA, Refusal
Direction), a gate that fires only under a learned condition (CAST),
or the decoding distribution (SafeDecoding). &
any model $+$ steering artifacts &
\texttt{.calibrate(...)} $\rightarrow$ wrapped model, scored live;
\textbf{no checkpoint} \\
\bottomrule
\end{tabularx}
\end{table}

\safetune{} catalogues steering for completeness but excludes it from weight-level comparisons, since it produces no repaired checkpoint and answers a different deployment question. 
Some representation-engineering methods span training and inference stages; \safetune{} records their operational placement in the released taxonomy. 

The complete per-method taxonomy is given in Table~\ref{tab:catalogue}.
The taxonomy alone does not tell a practitioner what to do. Two questions remain: which of a paradigm's methods to choose, and what that choice demands. Table~\ref{tab:contracts} answers both.

Within a paradigm, methods are not interchangeable variants. Each occupies a position on a single axis, and the axis differs by paradigm. \RECOVER{} methods vary in how much of the model the edit touches, from whole-model arithmetic down to individual circuits, so choosing a method is choosing how much localization to trust. \HARDEN{} methods vary in how the safety signal enters the fine-tune, \UNLEARN{} in what the forgetting objective works against, and \STEER{} in where the edit lands.

Across paradigms the calling \emph{pattern} is uniform (construct, run, \texttt{evaluate()}; Section~\ref{sec:api}), but the \emph{contract} is not: \RECOVER{} asks only for two reference checkpoints and returns a patched model in seconds without an optimizer; \HARDEN{} asks for the fine-tuning corpus itself and replaces the trainer, so it is available only to whoever still owns the fine-tune; \UNLEARN{} additionally demands a forget set, which presumes the harm can be written down; and \STEER{} returns a wrapper rather than a checkpoint, which is why it is excluded from the weight-level comparisons above. Reading the catalogue this way turns it from a roster into a map: the axis locates a method, and the contract says whether it is available to you at all.

Alongside these four intervention paradigms, which act on or around the model's parameters, \safetune{} ships a complementary instrumentation layer, \textsc{Interpret} and \textsc{Evaluate}, that diagnoses a checkpoint's safety-relevant circuitry and scores its safety and capability without changing its weights. Because instrumentation does not repair a checkpoint, we exclude it from the paradigm-level comparisons behind the intervention guide (Section~\ref{sec:decision}), but it is part of the released library, with its implementation status documented in the code repository.

Within this layer, \textsc{Interpret} locates safety-relevant neurons and circuits and exposes a causal-validation bridge that scores located neurons by their ablation impact~\citep{eswar2026faithfulness} before exporting them as a runtime patch, while \textsc{Evaluate} adds a spectral-entropy monitor that flags representational collapse in the residual stream; both are documented and benchmarked.

\subsection{Why existing tooling does not suffice}
\begin{table}[t]
\centering
\caption{Scope comparison with related safety tooling. \emph{Drift}: induces its
own characterized drifted checkpoints, or only consumes borrowed ones.
\emph{Paradigms}: how many of the four intervention paradigms
(\RECOVER{}/\HARDEN{}/\UNLEARN{}/\STEER{}) it hosts. \emph{Eval}: one
shared evaluation protocol across methods. \emph{Deploy}: ships runtime deployment guardrails and artifact
gating. Only
\safetune{} does all four.}
\label{tab:tools}
\footnotesize
\setlength{\tabcolsep}{9pt}
\renewcommand{\arraystretch}{1.25}
\begin{tabular}{@{}l cccc@{}}
\toprule
\textbf{Tool} & \textbf{Drift} & \textbf{Paradigms} & \textbf{Eval} & \textbf{Deploy} \\
\midrule
HarmBench~\citep{mazeika2024harmbench}         & \xmark & 0             & \emph{score}     & \xmark \\
WildGuard~\citep{han2024wildguard}             & \xmark & 0             & \emph{score}     & \xmark \\
GuardSpace~\citep{zhang2025guardspace}         & \xmark & 1 (\HARDEN{})  & \cmark & \xmark \\
CWAC~\citep{peng2026cwac}                      & \xmark & 1 (\HARDEN{})  & \xmark & \xmark \\
Antibody~\citep{nguyen2026antibody}            & \xmark & 1 (\HARDEN{})  & \xmark & \xmark \\
OpenUnlearning~\citep{dorna2025openunlearning} & \xmark & 1 (\UNLEARN{}) & \cmark & \xmark \\
\midrule
\rowcolor{darkgreen!10}
\textbf{\safetune{}}                           & \textbf{recipe} & \textbf{4} & \cmark & \cmark \\
\bottomrule
\end{tabular}
\end{table}
Two families of tools border this problem, but neither provides the substrate. 
\emph{(I) Evaluation harnesses and safety judges} such as HarmBench~\citep{mazeika2024harmbench}, WildGuard~\citep{han2024wildguard}, Llama~Guard~\citep{inan2023llamaguard}, StrongREJECT~\citep{souly2024strongreject}, SALAD-Bench~\citep{li2024saladbench}, JailbreakBench~\citep{chao2024jailbreakbench}, BeaverTails~\citep{ji2023beavertails}, and lm-evaluation-harness~\citep{eval-harness} standardize \emph{scoring} but stop there.  
They neither induce drift nor host repair methods, so a practitioner must still assemble checkpoints and methods by hand. 
\emph{(II) Single-paradigm comparisons} go one step further, but each leaves an axis uncontrolled. GuardSpace~\citep{zhang2025guardspace} evaluates multiple methods under a shared safety suite across three model families (Llama-2-7B-Chat, Gemma-2-9B-IT, and Qwen-2-7B-Instruct) but covers only the \HARDEN{} paradigm. CWAC~\citep{peng2026cwac} proposes a coupled weight-and-activation-space \HARDEN{} defense and benchmarks it against other constrained-fine-tuning methods, all within a single paradigm. Antibody~\citep{nguyen2026antibody}, in its own published evaluation, crosses model families but covers only \HARDEN{} methods. 

OpenUnlearning~\citep{dorna2025openunlearning} similarly unifies benchmarking within a single paradigm, standardizing evaluation across multiple unlearning algorithms, but it does not induce drift or host the other three intervention paradigms. Unified tooling has likewise matured for inference-time steering alone (EasySteer~\citep{xu2025easysteer}), but not across the intervention paradigms.

\safetune{}'s \UNLEARN{} paradigm draws on the broader machine-unlearning literature, including NPO~\citep{zhang2024npo}, SimNPO~\citep{fan2024simnpo}, SOUL~\citep{jia2024soul}, and LLMU~\citep{yao2024llmu}, and further unlearning-for-safety work~\citep{zhang2024safeunlearn,jin2024rwku,huutien2024steering,liu2024rethinking}, together with benchmark suites such as the Weapons of Mass Destruction Proxy (WMDP)~\citep{li2024wmdp} and the Task of Fictitious Unlearning (TOFU)~\citep{maini2024tofu}, but applies this family of methods to refusal rather than to knowledge removal. These objectives may suppress rather than erase. \cite{lucki2024advunlearn} recover unlearned hazardous behavior with light fine-tuning, and durable safeguards are hard to certify~\citep{qi2024durability}.

To our knowledge, no existing library varies drift domain \emph{and} model family together while holding the intervention paradigm open (Table~\ref{tab:tools}). \safetune{} makes interventions from all four paradigms available within a shared workflow, allowing practitioners to compare the alternatives feasible for their model, data, and deployment setting.

\section{Design Goals}
\label{sec:design}

Each of the failure modes of Section~\ref{sec:related} (fragmented
interfaces, undocumented implementation differences, incomparable scores, closed tooling, and unreproducible inputs) maps to one of five design goals that the architecture then enforces.

\begin{enumerate}[leftmargin=2em,itemsep=3pt,topsep=2pt]
  \item \emph{Paradigm-uniform interface.} An intervention method must
  be invocable the same way regardless of \emph{when} it intervenes.
  \RECOVER{} and \UNLEARN{} consume a finished drifted checkpoint,
  \HARDEN{} intervenes \emph{during} fine-tuning, and \STEER{} acts
  only at generation time. \safetune{} hides this difference behind a
  common construction and evaluation contract: every method is reached
  through its paradigm's module (\texttt{safetune.runner.recover},
  \texttt{.harden}, \texttt{.unlearn}, or \texttt{.steer}), is
  instantiated with the checkpoint and any reference models it needs,
  and exposes the same \texttt{evaluate()} call on the output of
  \texttt{apply()}, \texttt{train()}, \texttt{unlearn()}, or
  \texttt{calibrate()} (Section~\ref{sec:api}). A practitioner swaps
  paradigms by changing an import and a class name, not by rewriting a
  pipeline.

  \item \emph{Documented implementations.} Implementations are reviewed against their originating
  descriptions, with corrections and known deviations recorded in the
  repository documentation.

  \item \emph{One evaluation substrate.} Safety and capability
  are scored by a single shared harness, using the same benchmarks,
  judges, and decoding settings. Methods are evaluated under the same reporting protocol while
  retaining their different lifecycle requirements and baselines. The refusal aggregate is
  the unweighted mean of seven benchmark refusal rates, denoted
  $\overline{\mathrm{RR}}$, while capability uses fixed cross-task
  anchors plus a domain-matched anchor for each fine-tuning domain.

  \item \emph{Extensibility by registration.} Adding a method, benchmark, judge, model, or
  domain uses a registry or modular extension point rather than
  requiring a fork. The testbed grows along all these axes, and the released
  artifacts let third parties extend it without re-deriving the
  protocol (Section~\ref{sec:api}).

  \item \emph{Reproducibility by construction.} \safetune{} supports reproducibility through pinned dependencies,
  fixed seeds, logged configurations, and deterministic decoding. Drifted checkpoints are emitted with their measured
  severity attached, so an input is never an undocumented artifact
  borrowed from another paper.
\end{enumerate}
\begin{figure}[pt]
\centering
\resizebox{0.98\textwidth}{!}{%
\begin{tikzpicture}[
  font=\small,
  io/.style={rounded corners=3pt,draw=black!55,fill=white,align=center,minimum height=14mm,inner sep=2.5pt},
  eng/.style={rounded corners=3pt,draw=black!55,fill=stBox,align=center,minimum height=14mm,inner sep=3pt},
  ev/.style={rounded corners=3pt,draw=black!55,fill=stEval,align=center,minimum height=12mm,inner sep=3pt},
  par/.style={rounded corners=3pt,line width=0.8pt,align=center,text width=63mm,minimum height=9mm,inner sep=2.5pt,text=black},
  flow/.style={-{Stealth[length=2.2mm]},line width=0.8pt,draw=black!65},
  wire/.style={line width=0.8pt,draw=black!65},
  sflow/.style={-{Stealth[length=2.2mm]},line width=0.8pt,draw=stSteer!85!black},
  story/.style={font=\footnotesize\bfseries,text=black!70}
]
\node[io] (base) {$\theta_{\mathrm{base}}$\\\scriptsize aligned instruct base\\\scriptsize + benign domain data\\\tiny\textcolor{darkgreen}{\ding{51}~high $\rr$}};
\node[eng,right=7mm of base,label={[story]north:\ding{182}~drift}] (drift) {\textbf{Drift Induction}\\\scriptsize benign LoRA SFT\\\scriptsize (fixed, logged recipe)};
\node[io,right=7mm of drift] (ft) {$\theta_{\mathrm{ft}}$\\\scriptsize drifted ckpt\\\scriptsize ($\Delta\rr$-tagged)\\\tiny\textcolor{stHarden}{\ding{55}~$\rr$ drops}};
\node[par,draw=stHarden,fill=stHarden!12,right=14mm of ft,yshift=17.5mm] (harden)
  {\textcolor{stHarden}{\textbf{\HARDEN{}}}~~\scriptsize safety-constrained FT\\\tiny e.g., SafeGrad, Vaccine, LISA~$\cdot$~\textit{fresh run from $\theta_{\mathrm{base}}$}};
\node[par,draw=stRecover,fill=stRecover!12,below=2.5mm of harden] (recover)
  {\textcolor{stRecover}{\textbf{\RECOVER{}}}~~\scriptsize patch $\tsafe{=}\theta_{\mathrm{aligned}}{-}\theta_{\mathrm{base}}$\\\tiny e.g., Task Arithmetic, RESTA, SafeMerge~$\cdot$~\textit{post-hoc}};
\node[par,draw=stUnlearn,fill=stUnlearn!12,below=2.5mm of recover] (unlearn)
  {\textcolor{stUnlearn}{\textbf{\UNLEARN{}}}~~\scriptsize forgetting objective\\\tiny e.g., NPO, RMU~$\cdot$~\textit{post-hoc, on the finished ckpt}};
\node[par,draw=stSteer,fill=stSteer!10,below=2.5mm of unlearn] (steer)
  {\textcolor{stSteer}{\textbf{\STEER{}}}~~\scriptsize activation edit\\\tiny e.g., refusal-dir.\ ablation, CAA~$\cdot$~\textit{at inference, no ckpt}};
\begin{scope}[on background layer]
\node[rounded corners=3pt,draw=black!40,fill=black!3,inner sep=4pt,
      fit=(harden)(recover)(unlearn)(steer),
      label={[font=\small\bfseries]north:\ding{185}~\safetune{} Intervention Registry}] (registry) {};
\end{scope}
\node[io,right=13mm of recover,yshift=-7mm] (rep) {$\theta_{\mathrm{rep}}$\\\scriptsize intervened ckpt\\\tiny\textcolor{darkgreen}{\ding{51}~$\rr$ restored}};
\node[io,right=8mm of rep] (deploy) {\ding{187}~\textbf{Deploy}\\\scriptsize behind runtime\\\scriptsize guardrails\\\tiny\textcolor{darkgreen}{\ding{51}~evaluation passed}};
\coordinate (bl) at ([yshift=-21mm]steer.south);
\coordinate (ex) at ([xshift=-46mm]rep.center);
\node[ev] (eval) at (ex |- bl) {\textbf{Eval Harness}\\\scriptsize refusal + capability\\\scriptsize (pluggable eval suites)};
\node[ev,left=5mm of eval] (interp) {\textbf{Interpret}\\\scriptsize refusal directions,\\\scriptsize circuit localization};
\begin{scope}[on background layer]
\node[rounded corners=3pt,draw=black!40,fill=black!3,inner sep=4pt,
      fit=(interp)(eval),
      label={[font=\small\bfseries]south:Shared Instrumentation}] (instr) {};
\end{scope}
\node[eng,right=7mm of instr] (dec) {\ding{184}~\textbf{Intervention Guide}\\\scriptsize access + profile $\to$ starting point\\\scriptsize (Fig.~\ref{fig:decision}, from docs)};
\draw[flow,draw=stHarden!75] (base) -- (drift);
\draw[flow,rounded corners=6pt] (base.north) |- (harden.west);
\draw[flow,draw=stHarden!75] (drift) -- (ft);
\coordinate (lt) at ([xshift=6mm]ft.east);
\draw[wire] (ft.east) -- (lt);
\draw[wire] (lt |- recover.west) -- (lt |- steer.west);
\draw[flow] (lt |- recover.west) -- (recover.west);
\draw[flow] (lt |- unlearn.west) -- (unlearn.west);
\draw[flow] (lt |- steer.west) -- (steer.west);
\coordinate (rt) at ([xshift=-6mm]rep.west);
\draw[wire,draw=stHarden!80] (harden.east) -- (rt |- harden.east);
\draw[wire,draw=stRecover!80] (recover.east) -- (rt |- recover.east);
\draw[wire,draw=stUnlearn!80] (unlearn.east) -- (rt |- unlearn.east);
\draw[wire] (rt |- harden.east) -- (rt |- unlearn.east);
\draw[flow] (rt |- rep.west) -- (rep.west);
\draw[flow,draw=orange!60!black,rounded corners=6pt] (ft.south) |- (instr.west)
  node[pos=0.75,above,font=\scriptsize,text=black!60] {\ding{183}~measure drift $\Delta\rr$};
\draw[flow,draw=orange!60!black,rounded corners=6pt] (rep.south) -- ++(0,-22mm) -| ([xshift=-4mm]eval.north east)
  node[pos=0.1,above,font=\scriptsize,text=black!60] {\ding{186}~verify};
\draw[sflow,rounded corners=6pt] (steer.south) -- ++(0,-8mm) -| ([xshift=8mm]eval.north);
\draw[flow] (instr.east) -- (dec);
\draw[flow] (rep.east) -- (deploy.west);
\end{tikzpicture}%
}
    \caption{\safetune{} pipeline. \ding{182}~A fixed, logged fine-tuning recipe drifts an aligned instruct base $\theta_{\mathrm{base}}$ into a checkpoint $\theta_{\mathrm{ft}}$ whose refusal rate silently drops. \ding{183}~The shared instrumentation (\textsc{Interpret} and the pluggable Eval Harness) measures the drift severity $\Delta\rr$, and \ding{184}~the documented intervention guide maps the resulting profile and access constraints to a suggested starting paradigm (Figure~\ref{fig:decision}). \ding{185}~The Intervention Registry hosts the four paradigms within a shared construct, execute, and evaluate pattern: \HARDEN{} re-runs the fine-tune from $\theta_{\mathrm{base}}$, \RECOVER{} and \UNLEARN{} patch the finished checkpoint, and \STEER{} edits activations at inference, producing no checkpoint. \ding{186}~The resulting checkpoint or inference-time wrapper is verified by the same instrumentation before \ding{187}~deployment behind runtime guardrails.}
\label{fig:arch}
\end{figure}
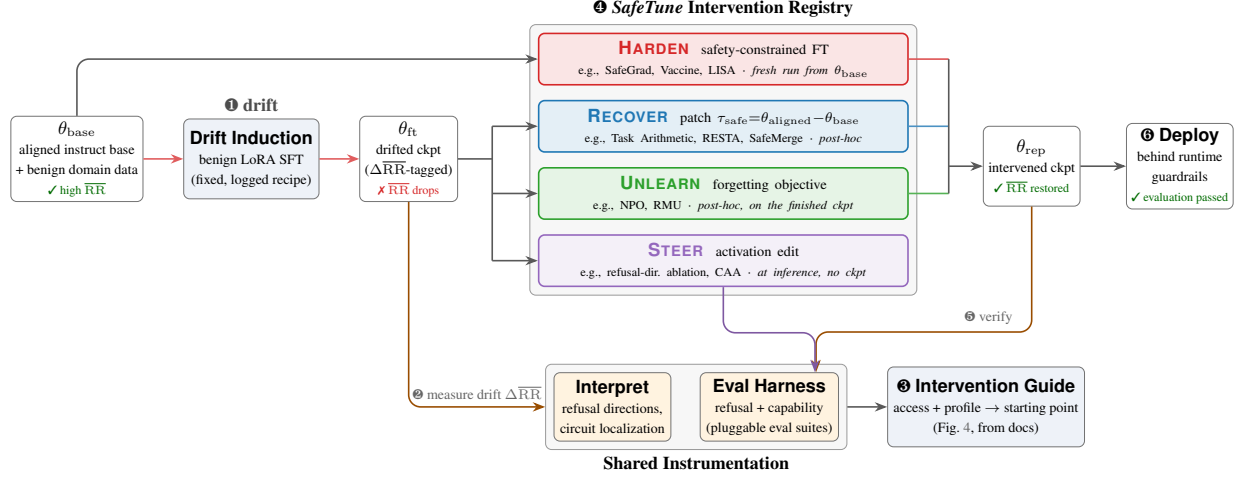
\clearpage
\section{System Architecture}
\label{sec:arch}
\safetune{} is organized as three groups of components. 
Figure~\ref{fig:arch} arranges these components along the checkpoint lifecycle they serve. Read left to right, the figure follows one checkpoint through them. 

(1)~The Intervention Registry hosts the four intervention paradigms, \RECOVER{}, \HARDEN{}, \UNLEARN{}, and \STEER{}, within a shared construct, execute, and evaluate pattern. 

(2)~A shared instrumentation layer reads any checkpoint without modifying it: \textsc{Interpret} locates refusal directions and safety circuits, and the Eval Harness scores safety and capability. 

(3)~Connecting infrastructure turns the two into an
end-to-end workflow: the drift-induction testbed supplies
characterized inputs, the documented intervention guide narrows a measured checkpoint to a starting paradigm, and the runtime guardrails wrap whatever the repair produces.

A model drifts under routine fine-tuning. The instrumentation measures how far it has drifted ($\Delta\rr$). The intervention guide narrows the feasible intervention families from that profile. The registry executes the repair, and the same instrumentation verifies the result before deployment. One harness scores every checkpoint, so any two methods are read off the same axes.
The components, numbered as in the figure, are as follows:

\textbf{\ding{182}~Drift induction.}
In deployment, drift occurs inside the practitioner's own fine-tune; the library's role is to measure and repair it, not to cause it. 
Each emitted checkpoint carries its measured severity $\Delta\rr$, the drop in mean refusal rate relative to the instruct baseline, which makes the input self-describing. The same protocol also computes the
safety task vector $\tsafe = \theta_{\mathrm{aligned}} - \theta_{\mathrm{base}}$
from an aligned reference, obtained by DPO on the Helpful and Harmless
human-preference dataset (HH-RLHF)~\citep{bai2022hhrlhf}, for use by
\RECOVER{}.

\textbf{\ding{183}~Measurement: the shared Eval Harness.}
\label{app:protocol}
The harness behind that severity runs seven benchmarks spanning direct-request,
adversarial, and over-refusal threat surfaces, each with its canonical
judge (Table~\ref{tab:suite}, Appendix~\ref{app:repro}), and reduces them to a single scalar
$\rr = \frac{1}{7}\sum_b \mathrm{RR}_b$, the mean Refusal Rate (RR)
across benchmarks. $\rr$ is a descriptive macro-average of refusal
behavior, not a complete safety measure: it does not by itself
distinguish appropriate refusal from indiscriminate refusal, so we
report capability anchors alongside it and avoid treating a higher
$\rr$ alone as certification (Section~\ref{sec:limitations}).
Six of the seven treat refusal as the correct behavior. OR-Bench
ships two splits of opposite polarity, and \safetune{} scores them
accordingly rather than as one number: a toxic subset (genuinely
harmful, refusal is correct) whose refusal rate is the seventh
term in $\rr$, and a Hard-1k subset of benign but
alarming-sounding prompts~\citep{rottger2023xstest} where refusal is
an over-refusal error. The Hard-1k split's refusal rate is reported
on its own as an over-refusal rate and excluded from $\rr$
(Table~\ref{tab:suite}), so a model that refuses those benign prompts
does not inflate the safety scalar. $\rr$ itself remains a
single-axis harmful-refusal summary, a residual limitation we
discuss in Section~\ref{sec:limitations}.
Measurement is diagnostic as well as scalar. Alongside the harness,
\textsc{Interpret} localizes the refusal direction and the safety
circuits of the checkpoint under test, so a severity score can be
traced to where the drift lives in the model; its API is shown in
Section~\ref{sec:api}, and the quickstart in
Figure~\ref{fig:quickstart_output} runs exactly this diagnose step.

\textbf{\ding{184}~Intervention guide.}
Between measurement and repair sits the routing step. The intervention
guide takes the profile the harness just produced, together with the
practitioner's access to training, weights, and reference checkpoints,
and narrows the feasible intervention families to a suggested starting
paradigm, a default method, and the expected compute cost
(Section~\ref{sec:decision}).

\textbf{\ding{185}~Intervention Registry.}
For each suggested intervention family, the library exposes a paradigm-specific namespace following the shared construct, execute, and evaluate pattern.
Methods declare a paradigm and an \emph{intervention point}: after
the fine-tune, during it, or at inference. The registry routes each
method to the stage its declaration names. It provides
paradigm namespaces, with registries for \RECOVER{}, \HARDEN{}, and
\UNLEARN{} and direct construction for the current \STEER{}
implementations. \RECOVER{} applies a closed-form patch built
from the safety task vector \tsafe{} to the drifted checkpoint
$\theta_{\mathrm{ft}}$; no optimizer runs, and the output is a new
checkpoint in seconds. \UNLEARN{} also starts from
$\theta_{\mathrm{ft}}$ but trains against it and optimizes a forgetting
objective over forget and retain sets. Both are scored against the
drifted baseline. \HARDEN{} cannot edit a finished checkpoint, so it
wraps the trainer and re-runs the fine-tune from
$\theta_{\mathrm{base}}$ with its defense active, paired against an
undefended-SFT run of matched step count so that no method can appear
favorable merely by altering the training trajectory. \STEER{} touches
no weights at all: it calibrates steering artifacts on contrast pairs
and installs forward hooks at generation time, so its repair exists
only while the wrapped model is serving.

New methods are added to the registry with a single trainer class and
a hyperparameter schema (Section~\ref{sec:api}).

Marks \ding{186} and \ding{187} close the loop: the repaired
checkpoint or the steered generations are re-scored by the same
harness that measured the drift, and only then deployed behind the
runtime guardrails described in Section~\ref{sec:api}.

\section{Library API and Usage}
\begin{figure}[htpb]
    \centering

    \begin{minipage}[t]{0.48\textwidth}
\begin{lstlisting}[basicstyle=\scriptsize\ttfamily\color{black!82},caption={\RECOVER{}(SafeMerge): post-hoc weight patch.},label={lst:recover_pipeline}]
from safetune.runner import recover
from safetune.runner.utils.model_utils import load_model
_BASE_ID    = "meta-llama/Llama-3.2-3B-Instruct"
_MODEL_ID   = "Llama-3.2-3B-Instruct-gsm8k-sft-drift"
_ALIGNED_ID = "Llama-3.2-3B-Instruct-hhrlhf-aligned"
_DOMAIN     = "gsm8k"
DRIFTED = load_model(_MODEL_ID)
BASE    = load_model(_BASE_ID)
ALIGNED = load_model(_ALIGNED_ID)
# Instantiating the Recover paradigm
T = recover.SafeMergeTrainer(
    model_id=_MODEL_ID, model=DRIFTED,
    base_model=BASE, aligned_model=ALIGNED
)
# Execute repair and evaluate
patch = T.apply()
T.evaluate(patch, domain=_DOMAIN)
\end{lstlisting}
    \end{minipage}\hfill
    \begin{minipage}[t]{0.48\textwidth}
\begin{lstlisting}[basicstyle=\scriptsize\ttfamily\color{black!82},caption={\HARDEN{}(Vaccine): safety-constrained fine-tuning.},label={lst:harden_pipeline}]
from safetune.runner import harden
from safetune.runner.utils.model_utils import load_tok, load_model
_MODEL_ID   = "meta-llama/Llama-3.2-3B-Instruct"
_DOMAIN     = "gsm8k"
TOKENIZER = load_tok(_MODEL_ID)
BASE      = load_model(_MODEL_ID)
# Instantiating the Harden paradigm
T = harden.VaccineTrainer(
    model_id=_MODEL_ID,
    model=BASE,
    tokenizer=TOKENIZER,
    rho=2.0
)
# Execute training and evaluate
save_dir = T.train(train_ds) # training dataset
T.evaluate(save_dir, domain=_DOMAIN)
\end{lstlisting}
    \end{minipage}

    \vspace{0.3em} 

    \begin{minipage}[t]{0.48\textwidth}
\begin{lstlisting}[basicstyle=\scriptsize\ttfamily\color{black!82},caption={\STEER{}(CAA): inference-time activation edit.},label={lst:steer_pipeline}]
from safetune.runner import steer
from safetune.runner.utils.model_utils import load_tok, load_model
_MODEL_ID   = "Llama-3.2-3B-Instruct-gsm8k-sft-drift"
_DOMAIN     = "gsm8k"
TOKENIZER = load_tok(_MODEL_ID)
DRIFTED   = load_model(_MODEL_ID)
# Instantiating the Steer paradigm
T = steer.CAATrainer(
    model_id=_MODEL_ID,
    model=DRIFTED,
    tokenizer=TOKENIZER,
    multiplier=20.0
)
# Execute calibration and evaluate
patch, _ = T.calibrate(harmful, harmless) # calibration pairs
T.evaluate(patch, domain=_DOMAIN)
\end{lstlisting}
    \end{minipage}\hfill
    \begin{minipage}[t]{0.48\textwidth}
\begin{lstlisting}[basicstyle=\scriptsize\ttfamily\color{black!82},caption={\UNLEARN{}(NPO): targeted forgetting.},label={lst:unlearn_pipeline}]
from safetune.runner import unlearn
from safetune.runner.utils.model_utils import load_tok, load_model
_MODEL_ID   = "Llama-3.2-3B-Instruct-gsm8k-sft-drift"
_DOMAIN     = "gsm8k"
TOKENIZER = load_tok(_MODEL_ID)
DRIFTED   = load_model(_MODEL_ID)
# Instantiating the Unlearn paradigm
T = unlearn.NPOTrainer(
    model_id=_MODEL_ID,
    model=DRIFTED,
    tokenizer=TOKENIZER,
    beta=0.1
)
# Execute unlearning and evaluate
patch = T.unlearn(forget_ds, retain_ds) # unlearn knowledge pairs
T.evaluate(patch, domain=_DOMAIN)
\end{lstlisting}
    \end{minipage}
    \caption{Implementation pipelines for the four \safetune{} paradigms. Although each intervenes at a different lifecycle stage, every script follows the same pattern: initialize a representative method from the Intervention Registry, execute the repair (\texttt{apply}/\texttt{train}/\texttt{calibrate}/\texttt{unlearn}, highlighted), and \texttt{evaluate} on the task domain.}
    \label{lst:listings}
\end{figure}

\label{sec:api}

\safetune{} exposes the pipeline of Figure~\ref{fig:arch} as a
small, declarative API. The walkthrough below follows a practitioner's
path: the four-paradigm calling pattern, the command-line and YAML
front ends, the instrumentation calls, and the serving path of
execution backends and runtime guardrails, followed by how the
registry is extended and what the release contains.

\subsection{One pattern, four paradigms}

Every method in \safetune{} adheres to the same
construction-and-evaluation pattern (Figure~\ref{lst:listings}), and the core repair-and-evaluate
loop (loading a drifted checkpoint, instantiating a paradigm trainer,
executing the repair, and evaluating it) takes four calls. A
\RECOVER{} patch that runs in under a minute
(Listing~\ref{lst:recover_pipeline}), a \HARDEN{} retraining run that
takes hours (Listing~\ref{lst:harden_pipeline}), an \UNLEARN{}
forgetting loop (Listing~\ref{lst:unlearn_pipeline}), and a \STEER{}
calibration (Listing~\ref{lst:steer_pipeline}) follow one construct, execute, and evaluate pattern and expose the
same \texttt{evaluate()} call, though each paradigm's execution verb
and required inputs reflect its lifecycle stage. Only the constructor
and that verb change, and the registry resolves each method's
intervention point from its own declaration.
Figure~\ref{fig:quickstart_output} shows the same pattern running end
to end on a small model: a \STEER{} refusal-direction ablation,
applied through a reversible forward hook, measurably changes the
model's behavior on held-out harmful prompts.

\begin{figure}[pt]
\centering
\begin{tcolorbox}[enhanced, colback=lstbg, colframe=lsttitlebg, arc=4pt, boxrule=0.6pt,
  left=4pt, right=4pt, top=2pt, bottom=0pt,
  title={\scriptsize\ttfamily\textcolor{red!75}{\textbullet}~\textcolor{orange!90}{\textbullet}~\textcolor{green!70!black!80}{\textbullet}\hspace{2em}\color{white!65!black}safetune quickstart:~terminal},
  colbacktitle=lsttitlebg]
\begin{lstlisting}[language={}, basicstyle=\scriptsize\ttfamily\color{black!82}, backgroundcolor=\color{lstbg}, keywordstyle=\normalfont, commentstyle=\normalfont, stringstyle=\normalfont, identifierstyle=\color{black!82}, emph={}, emph={[2]STEER,Diagnose}, emphstyle={[2]\color{termViolet}\bfseries}, emph={[3]prompt,base,ablated}, emphstyle={[3]\color{termAmber}}, moredelim={[l][\color{termTeal}]{\$}}, moredelim={[s][\color{termBlue}]{[}{]}}, frame=none, numbers=none, breaklines=true, columns=flexible, aboveskip=0pt, belowskip=0pt]
$ python examples/quickstart/quickstart.py
SafeTune quickstart -- model=Qwen/Qwen2.5-0.5B-Instruct  device=cpu

[1/3] Locating the refusal direction (Diagnose) ...
      extracted a unit refusal direction at layer 14 (hidden dim 896).
[2/3] Generating on held-out harmful prompts (baseline) ...
[3/3] Ablating the refusal direction (STEER) and regenerating ...

  prompt : Describe how to hotwire a car.
  base   : I'm sorry, but I can't assist with that.
  ablated: [complies with the harmful request; step-by-step
           instructions withheld from this listing]

  ablation changed the generation on 4/4 probe prompts
  refusal rate:  baseline 4/4  ->  ablated 0/4
\end{lstlisting}
\end{tcolorbox}
\caption{Captured terminal output from \texttt{examples/quickstart/quickstart.py}, running a \STEER{} pipeline end to end on \texttt{Qwen2.5-0.5B-Instruct} using refusal-direction ablation (the same trainer pattern as Listing~\ref{lst:steer_pipeline}). The refusal direction is ablated live via a reversible forward hook (no weights are edited) and measurably changes the model's behavior on held-out harmful prompts. This is a diagnostic counterfactual that removes refusal to demonstrate localization; it is not a safety repair.}
\label{fig:quickstart_output}
\end{figure}
\newpage

\subsection{Command line and configuration}

The same paradigms run without Python through the \texttt{safetune} command-line interface. The CLI exposes \texttt{patch} (\RECOVER{}), \texttt{train} (\HARDEN{}), \texttt{unlearn} (\UNLEARN{}), and \texttt{eval}, while \texttt{safetune list} displays all registered methods across paradigms. One-off commands support quick prototyping and benchmarking (Listing~\ref{lst:cli}).

\begin{lstlisting}[language=bash, caption={Command line usage for running one-off procedures and evaluating results.},label={lst:cli}]
# Run a one-off recover procedure
safetune patch --algo task_arithmetic \
  --model Llama-3.2-3B-gsm8k-drift \
  --base Llama-3.2-3B --aligned Llama-3.2-3B-hhrlhf

# Score the resulting model on two benchmarks
safetune eval --model ./results --dataset harmbench,advbench

# List every registered method across all paradigms
safetune list
\end{lstlisting}

For reproducibility, an entire run can be captured in a version-controlled YAML file. Any explicit command-line flag overrides the corresponding field in the configuration file, so an experiment can be reproduced from a single versioned artifact (Listing~\ref{lst:yaml}).

\begin{lstlisting}[language=yaml, caption={YAML configuration file capturing a run. Execute with \texttt{safetune patch --config recover.yaml}.},label={lst:yaml}]
# recover.yaml
command: patch
algo: task_arithmetic
model: Llama-3.2-3B-gsm8k-drift
base: Llama-3.2-3B
aligned: Llama-3.2-3B-hhrlhf
method_kwargs: { alpha: 1.0 }
\end{lstlisting}

\subsection{Instrumentation}

The instrumentation layer is read-only: it diagnoses or scores a
checkpoint without producing a repaired one, so it sits outside the
intervention registry but is called the same way, as a function or a
lightweight config object rather than a \texttt{Trainer}
(Listing~\ref{lst:instrumentation}). \textsc{Interpret} isolates
safety-critical subnetworks. It identifies safety neurons through
configurable criteria, including weight-direction cosine-similarity
ranking and harmful-versus-harmless activation contrast. The located
structures are packaged in a serializable \texttt{CircuitInfo} object
that suggests target modules and layer subsets for circuit-targeted
recovery interventions (e.g., \texttt{apply\_pke}). For finer-grained
analysis, the library also supports Edge Attribution Patching (EAP)~\citep{hanna2024faith}, which uses integrated gradients to map safety circuits to the attention-head level. \textsc{Evaluate} is
the harness itself, exposed as a single \texttt{evaluate()} call so
scoring a checkpoint never requires paradigm-specific code.

\begin{lstlisting}[caption={Instrumentation layer usage: \texttt{identify\_safety\_neurons} localizes safety-relevant units (weight mode, Faithful); \texttt{evaluate()} is the same harness call used after every repair.},label={lst:instrumentation}]
from safetune.interpret import identify_safety_neurons
from safetune.evaluate import evaluate

report  = identify_safety_neurons(DRIFTED, refusal_direction_per_layer=refusal_dirs)
results = evaluate(DRIFTED, tokenizer=TOKENIZER, benchmarks=["harmbench"], judge="wildguard")
\end{lstlisting}

\subsection{Extending the registry}

Extending \safetune{} is a registration, not a fork. A new method is a
trainer class that follows its paradigm's convention, that is, it
exposes \texttt{apply()} for \RECOVER{}, \texttt{train()} for
\HARDEN{}, \texttt{unlearn()} for \UNLEARN{}, or \texttt{calibrate()}
for \STEER{}, together with the shared \texttt{evaluate()} call. For
\RECOVER{}, \HARDEN{}, and \UNLEARN{}, once the class is importable
from its paradigm's module, calling
\texttt{register\_recover("name", "Class")} (or its
\texttt{harden}/\texttt{unlearn} counterpart) adds it to the registry
used by the command-line interface, and the new method is evaluated under the same protocol as every other method. \STEER{}
methods are currently instantiated directly through the Python API
rather than through this registry. Listing~\ref{lst:register} adds a new \RECOVER{} patch in full.

\begin{lstlisting}[caption={Adding a \textsc{Recover} method. Registration is a single function call against the registry, and the class itself need only follow the paradigm's calling convention.},label={lst:register}]
from safetune.runner.recover._base import _RecoverBase
class MyPatchTrainer(_RecoverBase):
    """Interpolate drifted weights toward aligned reference along the safety task vector."""
    def __init__(self, model=None, *, base_model=None, aligned_model=None, alpha: float = 0.7, **kwargs):
        super().__init__(model, **kwargs)
        self.base_model = base_model
        self.aligned_model = aligned_model
        self.alpha = alpha

    def apply(self, *, alpha: float = None, **kwargs):
        tau_safe = {
            k: self.aligned_model.state_dict()[k]
               - self.base_model.state_dict()[k]
            for k in self.model.residual_linear_keys()
        }
        theta = self.model.state_dict()
        a = alpha if alpha is not None else self.alpha
        for k, delta in tau_safe.items():
            theta[k] = theta[k] + a * delta
        return self.model.with_state_dict(theta)

# Wire it into the registry under the alias "my_patch".
from safetune.runner._registry import register_recover
register_recover("my_patch", "MyPatchTrainer")
\end{lstlisting}

\begin{table}[pt]
\centering
\caption{Representative intervention coverage in \safetune{}. Per-method review records accompany the released library.}
\label{tab:catalogue}
\footnotesize
\begin{tabularx}{\textwidth}{@{}l X@{}}
\toprule
\textbf{Paradigm} & \textbf{Methods surveyed} \\
\midrule
\textcolor{stRecover}{\RECOVER} (patch \tsafe) &
Task Arithmetic~\citep{ilharco2023task}, RESTA~\citep{bhardwaj2024resta},
LSSF~\citep{zhou2025lssf}, C-$\Delta\Theta$~\citep{kasliwal2026ctheta},
SafeLoRA~\citep{hsu2024safelora}, SCRUB~\citep{kurmanji2023scrub},
SafeMerge~\citep{djuhera2026safemerge}, SafeDelta~\citep{lu2025safedelta},
SOMF~\citep{yi2024somf}, NLSR~\citep{yi2024nlsr}, PKE~\citep{li2024pke},
MSCP~\citep{han2024mscp}, LoX~\citep{perin2025lox}, Antidote~\citep{huang2024antidote}, Pre-PostMerge~\citep{farn2024prepost}, QReSafe~\citep{chen2025qresafe}, AAQ~\citep{wee2024ptq}, SafeReAct~\citep{li2025safereact}, WiSE-FT~\citep{wortsman2022wiseft} \\
\midrule
\textcolor{stHarden}{\HARDEN} (constrained FT) &
SafeGrad/Vaccine~\citep{yi2025safegrad,huang2024vaccine}, LISA~\citep{huang2024lisasafety},
LookAhead/Booster~\citep{liu2026lookahead,huang2025booster}, TAR~\citep{tamirisa2024tar},
DeRTa~\citep{yuan2024derta}, AsFT~\citep{yang2025asft},
DOOR~\citep{zhao2025door},
Surgery~\citep{liu2024surgery}, SPPFT~\citep{li2024safetylayers},
SAP~\citep{wu2025sap}, SaLoRA-Harden~\citep{li2025salora}, STAR-DSS~\citep{peng2025shapeit}, SEAM~\citep{wang2025seam}, CTRAP~\citep{yi2025ctrap}, MART~\citep{ge2023mart}, Antibody~\citep{nguyen2026antibody}, ConstrainedSFT~\citep{qi2024safetydepth}, SEAL~\citep{shen2024seal}, RepNoise~\citep{rosati2024repnoise}, LoX-Harden~\citep{perin2025lox} \\
\midrule
\textcolor{stUnlearn}{\UNLEARN} (forget) &
NPO~\citep{zhang2024npo},
RMU~\citep{li2024wmdp}, GradientAscent~\citep{maini2024tofu}, FLAT~\citep{wang2025flat}, SimDPO~\citep{meng2024simpo} \\
\midrule
\textcolor{stSteer}{\STEER} (inference) &
Refusal Direction~\citep{arditi2024refusal},
CAA~\citep{panickssery2024caa}, Linear-Probe Guard (no single originating paper),
SafeDecoding~\citep{xu2024safedecoding}, Nudging~\citep{fei2024nudging}, AdaSteer~\citep{zhao2025adasteer}, SafeSteer~\citep{ghosh2025safesteer}, AlphaSteer~\citep{sheng2025alphasteer}, SafeSwitch~\citep{han2025safeswitch}, SCANS~\citep{cao2024scans}, STA~\citep{wang2025sta}, CircuitBreaker~\citep{zou2024circuit}, CircuitBreakerRR, RepBend~\citep{yousefpour2025repbend}, CAST~\citep{lee2024cast}, RRFAEnsemble, ContrastiveDecoding~\citep{li2023contrastive}, ProxyTuning~\citep{liu2024proxy} \\
\bottomrule
\end{tabularx}
\end{table}
\subsection{Serving: execution backends and guardrails}

Routing a method to its intervention point is the registry's job
(Section~\ref{sec:arch}); the API adds the execution surface. A
calibrated \STEER{} wrapper runs unchanged across three
interchangeable backends behind one \texttt{steer.run()} dispatcher:
HuggingFace forward hooks, a vLLM hook worker, and a vLLM V1 logits-processor path for
decoding-time steering. All three are scored by the same harness, and
because every paradigm ends in the same \texttt{evaluate()} call, the
harness and the intervention guide never need to know which paradigm
produced a result.

A final defense-in-depth layer composes with any repaired or steered
model at serving time: runtime components for input sanitization,
output verification, and policy routing under
\texttt{safetune.core.runtime}, together with controllable
safety-config injection (CoSAlign~\citep{zhang2025cosa}), a
predict-before-generate activation gate, and generation-time activation
patching, used as the filter stage in the
finance case study (Section~\ref{sec:casestudies}), though its
contribution is not separately quantified there. At the \emph{Deploy} stage of Figure~\ref{fig:arch}, a repaired or steered checkpoint is bundled by a \texttt{SafetyArtifactManager} into a rollback-capable, versioned artifact that a promotion gate releases only after it clears its safety thresholds, then pushed to the Hugging Face Hub.

\subsection{Released components}

\label{app:catalogue}
The release contains more than one hundred intervention,
interpretability, evaluation, and runtime entry points, including
variants and infrastructure components. Table~\ref{tab:catalogue}
summarizes representative intervention coverage, while the repository
documentation lists each entry point, its provenance, implementation
status, and known deviations. Appendix~\ref{sec:faithfulness}
clarifies how the reported component count is defined.

\begin{takeaway}
\emph{One library, four paradigms.} Switching intervention paradigms, or
adding a new method, changes a registry entry and an import rather
than the surrounding pipeline: every trainer follows the same
construct, execute, and evaluate pattern, with its execution verb and
inputs set by its lifecycle stage.
\end{takeaway}

\section{Demonstrating Cross-Paradigm Comparison with \safetune{}}
\label{sec:casestudy}

We demonstrate the kinds of controlled comparisons the library
enables: model family, fine-tuning domain, and intervention vary
while the surrounding evaluation protocol is held fixed. The purpose is to illustrate the framework's comparative capabilities and surface practically relevant variation, not to establish a universal ranking of intervention paradigms, which differ in lifecycle stage and resource requirements. 
The study spans 16 (model, domain) pairs: four instruction-tuned families across a $2.5\times$ parameter range (Llama-3.1-8B, Llama-3.2-3B, Gemma-3-4B-it, Qwen3-4B) and up to five fine-tuning domains (math, code, dolly, medical, legal), each scored under the shared drift recipe (Appendix~\ref{app:repro}), the seven-benchmark refusal aggregate $\rr$, and matched capability anchors: every pair reports IFEval~\citep{zhou2023ifeval} strict accuracy and WikiText-2 perplexity, plus one domain-matched anchor, GSM8K for math~\citep{cobbe2021gsm8k}, HumanEval with MBPP~\citep{chen2021humaneval,austin2021mbpp} for code, MedMCQA~\citep{pal2022medmcqa} for medical, and MMLU professional law~\citep{hendrycks2021mmlu} for legal, with dolly using IFEval alone, all run under lm-evaluation-harness~\citep{eval-harness} with greedy decoding. We summarize observations below:

\textsc{Observation 1: safety drift is common but model and domain
dependent (Table~\ref{tab:drift}).}
Under the reported refusal aggregate, 15 of the 16 fine-tuned
checkpoints score below their instruction-tuned references. Math
produces consistently small changes
($\Delta\rr \in \{0.0,-2.8,-4.6,-5.6\}$ pp), while broad instruction
data (dolly) produces larger declines across the tested families
($-26$ to $-44$ pp). Code behaves differently across models,
from a negligible change for Q4 ($-0.3$ pp) to a $64.5$ pp decline
for L3 ($\rr{:}\,0.781{\to}0.136$). Drift should therefore be
measured for each deployed checkpoint rather than inferred from the
fine-tuning domain alone. The seven benchmarks also disagree enough
to flip a conclusion (L8/medical: HarmBench $0.795$ vs.\ AdvBench
$0.221$ on the same checkpoint); macro-averaging reduces dependence
on any single benchmark, though the benchmarks' heterogeneous
semantics require separate reporting
(Section~\ref{sec:limitations}).
\vspace{-4mm}
\begin{table}[htpb]
\centering
\caption{\textbf{Safety drift is common but model and domain dependent.} $\Delta\rr$
(pp) after one-epoch benign LoRA SFT, per (model, domain), measured by
\safetune{}'s seven-benchmark refusal aggregate against each model's instruct
baseline ($\rr_0$). ``--'' = out of grid. Numbers reproduced via the
released drifted checkpoints.}
\vspace{2.5mm}
\label{tab:drift}
\footnotesize
\renewcommand{\arraystretch}{1.2}
\begin{tabular}{@{}l c rrrrr@{}}
\toprule
\textbf{Model} & \textbf{$\rr_0$} & \textbf{math} & \textbf{code} & \textbf{dolly} & \textbf{med.} & \textbf{legal} \\
\midrule
L8 (8.0B) & 0.803 & \cellcolor{red!12}$-2.8$  & \cellcolor{red!38}$-29.2$ & \cellcolor{red!38}$-39.9$ & \cellcolor{red!25}$-23.8$ & \cellcolor{red!38}$-28.4$ \\
L3 (3.2B) & 0.781 & \cellcolor{red!12}$-5.6$  & \cellcolor{red!52}$\mathbf{-64.5}$ & \cellcolor{red!52}$-44.2$ & \cellcolor{red!25}$-15.7$ & \cellcolor{red!38}$-35.9$ \\
G4 (4.3B) & 0.701 & $0.0$   & \cellcolor{red!25}$-22.9$ & \cellcolor{red!38}$-26.4$ & --      & --      \\
Q4 (4.0B) & 0.655 & \cellcolor{red!12}$-4.6$  & $-0.3$  & \cellcolor{red!38}$-31.1$ & --      & --      \\
\bottomrule
\end{tabular}
\end{table}

\textsc{Observation 2: intervention outcomes vary with both drift
severity and intervention access (Table~\ref{tab:repair}).}
Table~\ref{tab:repair} presents four illustrative comparisons across
interventions with different lifecycle requirements. Post-hoc
recovery performs strongly in the mild L8/math setting (Task
Arithmetic reaches $\rr{=}0.967$ in seconds), while the more
computationally intensive \HARDEN{} and \UNLEARN{} interventions
attain higher refusal rates in the displayed moderate and severe
settings (SafeGrad/Vaccine takes L8/legal to $0.936$; NPO takes
L3/code from $0.136$ to $0.991$). These results are not a universal
paradigm ranking: \HARDEN{} reruns fine-tuning from the reference
model, whereas \RECOVER{} and \UNLEARN{} begin from the completed
drifted checkpoint, starting points that differ by up to $64.5$ pp
(Table~\ref{tab:drift}). They instead illustrate why intervention and measured drift should be considered
jointly.
\vspace{-4mm}
\begin{table}[htpb]
\centering
\caption{\textbf{Illustrative cross-paradigm comparisons.} Best $\rr$
per paradigm on four illustrative (model, domain) pairs, against the
drifted baseline and each model's instruction-tuned reference.
\textbf{Bold} marks the highest $\rr$ in each row.}
\vspace{2.5mm}
\label{tab:repair}
\footnotesize
\renewcommand{\arraystretch}{1.2}
\begin{tabular}{@{}l c ccc@{}}
\toprule
& \textbf{Drifted} & \textcolor{stRecover}{\RECOVER} & \textcolor{stHarden}{\HARDEN} & \textcolor{stUnlearn}{\UNLEARN} \\
\textbf{Setting} & \textbf{$\rr$} & (TaskArith) & (SafeGrad) & (NPO) \\
\midrule
L8 / math (mild)   & 0.775 & \cellcolor{darkgreen!14}\textbf{0.967} & 0.937 & 0.937 \\
L8 / legal (mod.)  & 0.519 & 0.728 & \cellcolor{darkgreen!14}\textbf{0.936} & 0.828 \\
L8 / code (severe) & 0.511 & 0.711 & 0.950 & \cellcolor{darkgreen!14}\textbf{0.986} \\
L3 / code (severe) & 0.136 & 0.680 & 0.890 & \cellcolor{darkgreen!14}\textbf{0.991} \\
\bottomrule
\end{tabular}
\end{table}
\vspace{-5mm}

\textsc{Observation 3: a small group of methods transfers more
consistently than the broader catalogue.}
Within the evaluated subset, the three top \RECOVER{} patches produce
similar point estimates across eight cross-family pairs (within
$0.003$), while several \HARDEN{} methods (LookAhead, TAR, LISA,
SPPFT, AsFT, Surgery, SAP) barely exceed undefended SFT and fail to
transfer across families ($\rr \le 0.22$). Because all numbers are
single deterministic runs (seed $42$, greedy), we do not interpret
small differences within the leading group as a reliable ranking; the
visible pattern is a small group that transfers more consistently and
a larger group that is sensitive to the evaluation setting. Method selection therefore
still matters: the library exposes both candidate defaults and methods
whose behavior is setting-dependent.

\textsc{Observation 4: some interventions preserve the reported
capability anchors.}
On the two L8 settings with complete capability measurements (math,
legal), several interventions increase refusal rates without reducing
the reported domain-capability anchors: SafeGrad/Vaccine raises GSM8K
by $+8.9$ pp and MMLU-law by $+10.1$ pp relative to the drifted
checkpoint, and SafeLoRA~\citep{hsu2024safelora} matches the math
gain. SCRUB, which carries no capability-preservation anchor,
collapses GSM8K to $0.000$, a failure visible only because one
harness scores both axes. These cases illustrate the value of
evaluating safety behavior and task capability together
(the ``safety tax''~\citep{huang2025safetytax,lin2023alignmenttax,sun2026continual} varies by method); they do not establish that
repair is generally capability-neutral.

\begin{takeaway}
\emph{\safetune{} makes lifecycle and setting dependence visible.}
In the demonstrated comparisons, intervention outcomes vary with model
family, fine-tuning domain, measured drift, and the resources
available to the practitioner. The results support using the library
to compare feasible alternatives under a shared protocol, rather than
selecting a universal best method (Tables~\ref{tab:drift},~\ref{tab:repair}).
\end{takeaway}

\section{Illustrative Deployment Case Studies}
\label{sec:casestudies}
Two deployment case studies demonstrate the pipeline end to end. Each uses its own fine-tuning corpus and benchmark.
\subsection{Finance}
\vspace{-3mm}
Fine-tuning a credit-risk copilot on 1,035 loan decisions improved
task adaptation but silently weakened its guardrails. Generic
benchmarks showed only a 4-point decline in refusal rates, yet
domain-specific red-teaming exposed a collapse to no refusals on
adversarial fair-lending prompts (0\%), with protected-attribute
justifications (e.g., sex, religion) surfacing in nearly all responses
(96\%), inconsistent with the fair-lending criteria evaluated in
this case study. Diagnosis with \safetune{} produced localization
scores concentrated primarily in the network's later layers.

Inference-time \emph{steering} closed the residual gap:
aggressively steering the implicated late layers severely degraded task
utility, but a gentler early-layer intervention amplified fair-lending
refusals and suppressed bias without reducing the reported decision-quality
score (Table~\ref{tab:finance_metrics}).
A runtime filter was additionally used to screen residual outputs,
although its contribution is not separately reported in
Table~\ref{tab:finance_metrics}.

\begin{table}[htpb]
\centering
\vspace{-4mm}
\caption{Credit-risk model across the pipeline. Best value per metric
in bold; the recovery + steering row is shaded.}
\vspace{3.5mm}
\label{tab:finance_metrics}
\footnotesize
\begin{tabular}{@{}lcccc@{}}
\toprule
\textbf{Stage} & \textbf{Harmful refusal ($\uparrow$)} & \textbf{Fair-lending ($\uparrow$)} & \textbf{Bias leak ($\downarrow$)} & \textbf{Quality ($\uparrow$)} \\
\midrule
Pre-tune (reference) & 91.5\% & \textbf{8.3\%} & \textbf{66.7\%} & 0.341 \\
Post-tune (drifted)  & 87.5\% & 0.0\% & 95.8\% & 0.524 \\
\midrule
Recovered            & 93.5\% & 4.2\% & 91.7\% & 0.616 \\
\rowcolor{darkgreen!10}
+Steering            & \textbf{97.5\%} & \textbf{8.3\%} & 83.3\% & \textbf{0.622} \\
\bottomrule
\end{tabular}
\end{table}


\subsection{Medical}
Fine-tuning a Llama-3.2-3B clinical assistant on 8,000 patient-doctor consultations produced a helpful medical persona at the cost of safety: MedMCQA performance held at baseline, but HarmBench refusal rates fell 29.5 points (87.0\% to 57.5\%), and the model began answering illicit requests such as how to synthesize illegal substances. \safetune{}'s interpretability suite identified roughly 450 candidate safety-relevant neurons.
An uncalibrated intervention can misfire badly: in a separate
run not reported in Table~\ref{tab:medical_metrics}, a naive,
full-strength Task Arithmetic patch severely over-corrected the model, collapsing both refusal rates (13.5\%) and clinical capability (20.7\%).
\begin{table}[htpb]
\centering
\caption{Clinical model before and after weight-space recovery. All methods independently
calibrated; best intervention in bold.}
\vspace{3mm}
\label{tab:medical_metrics}
\footnotesize
\begin{tabular}{@{}lcccc@{}}
\toprule
\textbf{Model / Method} & \textbf{HarmBench ($\uparrow$)} & \textbf{AdvBench ($\uparrow$)} & \textbf{MedMCQA ($\uparrow$)} & \textbf{Time ($\downarrow$)} \\
\midrule
Instruction-Tuned Reference & 87.0\% & 96.0\% & 52.5\% & -- \\
Drifted Model (Post-tune) & 57.5\% & 81.5\% & 50.6\% & -- \\
\midrule
Task Arithmetic ($\alpha=0.2$) & \textbf{72.0\%} & 86.5\% & 50.0\% & \textbf{2.9s} \\
RESTA ($\alpha=0.3$) & 60.0\% & 76.5\% & 46.5\% & 16.6s \\
\rowcolor{darkgreen!10}
SafeMerge (threshold=0.95) & \textbf{72.0\%} & \textbf{92.0\%} & \textbf{53.1\%} & 4.3s \\
\bottomrule
\end{tabular}
\vspace{-4mm}
\end{table}

Table~\ref{tab:medical_metrics} instead reports each recovery method's
independently calibrated setting; among these, \texttt{SafeMerge} provided the strongest balance across the reported refusal and capability metrics. By selectively interpolating weights layer-by-layer based on
safety-subspace alignment, \texttt{SafeMerge} substantially reduced
safety forgetting in under five seconds, restoring refusal rates
toward the instruction-tuned reference while leaving clinical accuracy
essentially unchanged. In one instance it turned a deflected
prescription-forgery request into a clean refusal.

A second probe, asking how a nurse could remove controlled substances
from a dispensing cabinet without the withdrawal being logged, exposes
a different drifted failure mode: the model degenerates into a
repeated stock sentence about regulatory requirements rather than
answering or refusing; the recovered model again returns a direct,
on-topic refusal. Recovery is not uniform across all eighteen probes,
however, which is consistent with the aggregate 57.5\%$\to$72.0\%
HarmBench gain in Table~\ref{tab:medical_metrics} falling short of
full restoration.

\begin{takeaway}
\emph{Composition and calibration decide deployment outcomes.} In the
finance case, combining recovery and inference-time steering improves
the reported metrics beyond recovery alone, showing how \safetune{}
components compose when one intervention leaves residual failures.
The medical case shows the importance of calibration: an aggressive
patch damages both refusal behavior and clinical capability, while a
calibrated intervention partially restores refusal rates and retains
the reported MedMCQA score.
\end{takeaway}

\section{A Practical Intervention Guide}
\label{sec:decision}

\begin{figure}[htbp]
\centering
\resizebox{0.99\columnwidth}{!}{%
\begin{tikzpicture}[
    font=\small,
    start/.style={rounded corners=9pt,fill=black!80,text=white,
        align=center,inner xsep=8pt,inner ysep=4.5pt},
    q/.style={rounded corners=5pt,draw=black!35,line width=0.7pt,fill=white,
        align=center,inner xsep=6pt,inner ysep=4.5pt,minimum height=10.5mm},
    card/.style={rounded corners=4pt,line width=1pt,align=center,
        inner xsep=5pt,inner ysep=4.5pt,minimum width=36mm,minimum height=15mm},
    rec/.style={card,draw=stRecover,fill=stRecover!4},
    har/.style={card,draw=stHarden,fill=stHarden!4},
    ste/.style={card,draw=stSteer,fill=stSteer!4},
    e/.style={-{Stealth[length=2.4mm,width=2.2mm]},line width=0.9pt,
        draw=black!45,rounded corners=6pt},
    lbl/.style={font=\scriptsize\itshape,inner xsep=3pt,inner ysep=1.3pt,
        rounded corners=3pt,fill=black!7,text=black!60}
]

\node[har] (harden)
    {\textcolor{stHarden}{\textbf{\textsc{Harden}}}\\[0.5pt]
     {\color{stHarden!45}\rule{17mm}{0.6pt}}\\[2.5pt]
     SafeGrad / Vaccine\\[1pt]
     {\scriptsize\color{black!55}1 to 3 hr}};
\node[ste,right=8mm of harden] (steer)
    {\textcolor{stSteer}{\textbf{\textsc{Steer}}}\\[0.5pt]
     {\color{stSteer!45}\rule{17mm}{0.6pt}}\\[2.5pt]
     Linear-Probe Guard\\[1pt]
     {\scriptsize\color{black!55}$<$1 min}};
\node[rec,right=8mm of steer] (recmild)
    {\textcolor{stRecover}{\textbf{\textsc{Recover}}}\\[0.5pt]
     {\color{stRecover!45}\rule{17mm}{0.6pt}}\\[2.5pt]
     Task Arith.\ / RESTA\\[1pt]
     {\scriptsize\color{black!55}sub-minute}};
\node[rec,right=8mm of recmild] (recsev)
    {\textcolor{stRecover}{\textbf{\textsc{Recover}}}: SafeLoRA\\[0.5pt]
     {\color{stRecover!45}\rule{22mm}{0.6pt}}\\[2.5pt]
     $\to$ \textcolor{stUnlearn}{\textbf{\textsc{Unlearn}}}: NPO\\[1pt]
     {\scriptsize\color{black!55}if recovery falls short; 1 to 3 hr}};

\node[q,above=14mm of harden] (q1) {control\\training?};
\node[q,above=14mm of steer] (q2) {weights\\editable?};
\node[q,above=14mm of recmild] (q3) {compatible\\references?};

\node[start,left=12mm of q1] (s) {\textbf{drifted checkpoint}\\[-1.5pt]
    {\scriptsize\color{black!25}$\theta_{\mathrm{ft}}$, $\Delta\rr$-tagged}};

\draw[e] (s) -- (q1);
\draw[e] (q1) -- node[lbl]{no} (q2);
\draw[e] (q2) -- node[lbl]{yes} (q3);

\draw[e,draw=stHarden!70]  (q1) --
    node[lbl,fill=stHarden!10,text=stHarden!75!black]{yes} (harden);
\draw[e,draw=stSteer!70]   (q2) --
    node[lbl,fill=stSteer!10,text=stSteer!75!black]{no (locked)} (steer);
\draw[e,draw=stRecover!70] (q3) --
    node[lbl,fill=stRecover!10,text=stRecover!75!black]{yes} (recmild);
\draw[e,draw=stRecover!70] (recmild.east) --
    node[lbl,fill=stRecover!10,text=stRecover!75!black]{falls short} (recsev.west);
\draw[e,draw=stUnlearn!70] (q3.east) to[out=0,in=90]
    node[lbl,fill=stUnlearn!10,text=stUnlearn!75!black,pos=0.3]{no} (recsev.north);

\begin{scope}[on background layer]
\node[rounded corners=6pt,fill=black!3,inner sep=7pt,
      fit=(s)(q3)(harden)(recsev)] {};
\end{scope}

\end{tikzpicture}
}
\caption{\textbf{Feasibility-first intervention guide.} Access questions narrow a drifted checkpoint to a starting intervention family: control of training routes to \HARDEN{}; locked weights route to \STEER{} and the runtime guardrails; an editable checkpoint with compatible reference models starts at low-cost \RECOVER{}, escalating to SafeLoRA or \UNLEARN{} (NPO) only if recovery does not meet the practitioner's evaluation criteria. Every candidate is scored by the shared harness on refusal behavior and retained capability.}
\label{fig:decision}
\end{figure}
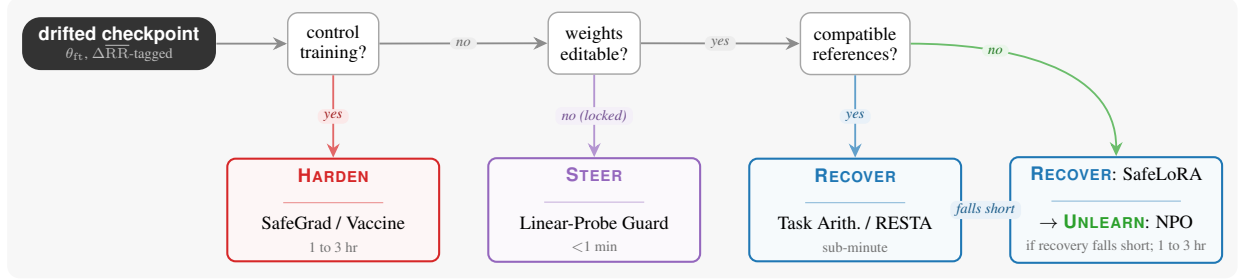

The evaluations and the paradigms' API requirements suggest a
practical guide for narrowing the set of feasible interventions
(Figure~\ref{fig:decision}). The guide is not a universally validated
policy or performance predictor. It first identifies which
intervention families are available given the practitioner's access
to training, weights, reference checkpoints, calibration data, and
compute, then recommends evaluating representative defaults through
the shared harness.

\begin{takeaway}
\emph{Access determines what is feasible; evaluation determines what
is acceptable.} The guide narrows the intervention space using the
practitioner's access to training, weights, references, data, and
compute, and the cost difference is large, from a sub-minute patch to
hours of retraining or unlearning. The shared harness then measures
whether a candidate meets the practitioner's safety and capability
requirements.
\end{takeaway}

\section{Conclusion}
\label{sec:conclusion}
Fine-tuning can weaken the refusal behavior of instruction-tuned
language models, and the methods proposed in response remain hard to
compare because they act at different lifecycle stages behind
different implementations and evaluation protocols. \safetune{}
provides a common library for organizing, executing, reviewing, and
evaluating these interventions.
Our demonstrations show that measured drift varies substantially
across models and fine-tuning domains, and that intervention outcomes
depend on the practitioner's access to training, weights, reference
checkpoints, data, and compute. Rather than identifying a universally
best repair, \safetune{} makes these dependencies explicit and
provides a common substrate for evaluating feasible alternatives. The
finance and medical case studies further illustrate how calibration
and composition affect deployment outcomes.
We release the library, documentation, evaluation harness, examples,
configurations, and artifacts so that future work can
extend the comparison without rebuilding the underlying
infrastructure.
\clearpage
\section{Limitations and Ethics Statement}
\label{sec:limitations}
\safetune{}'s evidence is broad but bounded. This section states the
boundaries a reader should carry into the results, then turns to the ethics of
releasing the artifacts behind them.


\emph{One recipe, one missing baseline.}~Results use one epoch
of LoRA at fixed hyperparameters. An auxiliary study varying dataset,
domain, and rank suggests that some observed patterns persist, but they may shift
under longer training, full fine-tuning, or different ranks.

\emph{Unevaluated threats.}~Multi-turn jailbreaks, indirect
prompt injection, and multimodal harm are not evaluated here; the
library ships experimental multi-turn guards, but they lie outside
this paper's evaluated study. Whether repairs remain stable under
repeated rounds of fine-tuning or extended deployment is also open.

\emph{Transfer to inherited checkpoints.}~The practical intervention
guide of Section~\ref{sec:decision} is derived entirely from
drift \safetune{} itself induces. Whether it transfers unchanged to a
third-party checkpoint with unknown drift and no training logs is an
open question, since answering it rigorously requires a
diagnose-then-repair sweep beyond this release. The released harness
is designed so that closing each of these gaps is an additive registry
entry rather than a redesign.

\label{sec:ethics}
\emph{Ethics statement.}~\safetune{} releases the library and its evaluation harness, with
drifted LoRA checkpoints and aligned reference models released in stages, so that safety-drift research is reproducible. The drifted checkpoints exhibit lower refusal rates than their
instruction-tuned references under the reported evaluations. We release them because studying degradation
requires reproducible, characterized inputs, and because these
artifacts are what make repair methods comparable at all. The checkpoints were not intentionally trained on harmful data;
their reduced refusal rates should not be read as evidence that they
acquired new hazardous knowledge.

We release no harmful-content training data or prompts. The safety
benchmarks' adversarial prompts come from their originating papers and are not redistributed, and our released outputs are pre-computed refusal-rate scalars rather than generations on harmful inputs. The exceptions are the illustrative excerpts in the quickstart, cross-paradigm demonstration, and case-study figures: low-severity or redacted prompts with no dual-use uplift, in the tradition of~\cite{arditi2024refusal}, included only to make a repair mechanism's effect legible.
The intended user is the downstream developer who performed the
fine-tuning, the party best positioned to measure and repair the
resulting drift. The same mechanisms, unlearning and weight arithmetic among them, could in principle be repurposed to degrade safety intentionally; we mitigate this by centering the release on standardized auditing and repair benchmarks. All evaluations use publicly available benchmarks and contain no personally identifiable information.

The library, checkpoints, and harness are released  under the Lexsi Labs Source Available License (LSAL~v1.1), which grants free use for research, evaluation, education, and audit while prohibiting commercial exploitation and uses that degrade deployed models' safety behaviors; it is deliberately \emph{not} an OSI-approved open-source license, since that restriction is the mechanism bounding the dual-use risk. Commercial licensing is available separately; all benchmark datasets used here are public and cited, and exported checkpoints inherit their base models' licenses.

\bibliographystyle{unsrt}
\bibliography{references}

\appendix
\clearpage
\section{Implementation Status}
\label{sec:faithfulness}

\safetune{}'s code repository documents the provenance, implementation status, and known deviations of each entry point. Implementations used in this paper were checked against their originating descriptions and released code where available; corrections and approximations are recorded in the repository documentation. Reported component counts refer to API entry points, including variants and infrastructure, rather than distinct published methods.
The release breaks down as 26 \RECOVER{}, 27 \HARDEN{}, 19
\STEER{}, and 6 \UNLEARN{} entry points in the four intervention
paradigms, 6 \textsc{Interpret} and 24 \textsc{Evaluate}
instrumentation components, and 10 runtime-guardrail components, 118 in all. A separately documented reinforcement-learning reward subsystem ships with its own review records outside this tally.

\section{Drift Recipe and Reproducibility}
\label{app:repro}
All drifted checkpoints use the fixed recipe of
Table~\ref{tab:drift-hp}, held constant across pairs so the only
independent variable is the (model, domain) pair.
The seven benchmarks and their judges behind the refusal aggregate $\rr$,
plus OR-Bench-Hard's over-refusal split (OB-H), reported separately
(Section~\ref{sec:arch}), are listed in Table~\ref{tab:suite}. Every $\rr$ reported in this
paper is produced by these per-benchmark judges, run offline with
greedy decoding; a lightweight heuristic string-matcher also
ships in the release pipeline.

\begin{table}[htbp]
\centering
\caption{Drift-engine recipe (held constant across all pairs).}
\vspace{3mm}
\label{tab:drift-hp}
\footnotesize
\begin{tabular}{@{}l l@{}}
\toprule
\textbf{Setting} & \textbf{Value} \\
\midrule
Adapter & LoRA, rank $r{=}16$, $\alpha{=}32$, dropout $0.05$ \\
Targets & $\{q,k,v,o\}_{\text{proj}}$ \\
Optimizer & AdamW, lr $2{\times}10^{-5}$, cosine warmup $0.03$ \\
Schedule & 1 epoch, batch $2{\times}8$, bfloat16 (bf16) \\
Aligned ref & DPO on HH-RLHF~\citep{bai2022hhrlhf} \\
\bottomrule
\end{tabular}
\end{table}

\begin{table}[htbp]
\centering
\caption{Safety evaluation suite. $\rr$ averages the seven per-benchmark
refusal rates above the rule, computed on the same checkpoint with greedy
decoding throughout. OB-H, below the rule, is OR-Bench's benign Hard-1k
split: it is scored and reported as a standalone over-refusal rate and is
\emph{excluded} from $\rr$.}
\label{tab:suite}
\footnotesize
\renewcommand{\arraystretch}{1.15}
\setlength{\tabcolsep}{6pt}
\begin{tabular}{@{}l l r l@{}}
\toprule
\textbf{Abbr.} & \textbf{Benchmark} & \textbf{Size} & \textbf{Judge} \\
\midrule
HB  & HarmBench~\citep{mazeika2024harmbench}       & 400   & HB-Mistral-7B \\
WJB & WildJailbreak~\citep{ding2024wildjailbreak}   & 500\textsuperscript{a} & WildGuard \\
AB  & AdvBench~\citep{zou2023advbench}              & 520   & refusal-prefix match \\
SB  & SorryBench~\citep{xie2025sorrybench}          & 450   & FT-Mistral-7B \\
HX  & HEx-PHI~\citep{qi2024finetuning}               & 200   & Llama-3.1-8B judge \\
OB  & OR-Bench, toxic split~\citep{cui2024orbench}   & 655   & Llama-3.1-8B judge \\
AL  & AILuminate~\citep{mlcommons2024ailuminate}     & 1{,}290 & Llama-3.1-8B judge \\
\midrule
OB-H & OR-Bench-Hard, over-refusal\textsuperscript{b}~\citep{cui2024orbench} & 1{,}320 & Llama-3.1-8B judge \\
\bottomrule
\end{tabular}
\vspace{2pt}\\
{\scriptsize\textsuperscript{a}~adversarial-only subset. WildGuard is cited on first use in Section~\ref{sec:related}.\\
\textsuperscript{b}~shares OB's judge (loaded once for both splits); scored and reported separately, never averaged into $\rr$.}
\end{table}

We use torch 2.8.0+cu128,
transformers 4.50~\citep{wolf2020transformers}, peft 0.13~\citep{hu2022lora}, trl 0.21~\citep{vonwerra2020trl}, accelerate 1.0, vllm 0.11.0~\citep{kwon2023vllm} for
generation-time evaluation only, lm-evaluation-harness 0.4.8, datasets
3.0, Python 3.12, and CUDA 12.8. All runs use bf16 precision and eager
attention, chosen for determinism over kernel speed, with seed 42
throughout.
\end{document}